%% file: main.tex
\documentclass[11pt]{article}
\usepackage[utf8]{inputenc}
\usepackage[T1]{fontenc}
\usepackage{lmodern}
\usepackage[margin=1in]{geometry}
\usepackage{amsmath,amssymb,amsthm}
\usepackage{graphicx}
\usepackage{booktabs}
\usepackage{xcolor}
\usepackage{microtype}
\usepackage{caption}
\usepackage{float}
\usepackage[section]{placeins}
\usepackage[hidelinks]{hyperref}
\usepackage{url}
\newtheorem{proposition}{Proposition}

\providecommand{\Rspear}{-0.00}
\providecommand{\RspearBest}{-0.00}
\providecommand{\Rntrain}{0}
\providecommand{\Rntest}{0}

\providecommand{\Rnseries}{0}
\providecommand{\Rnwin}{0}

\providecommand{\RoursAcc}{0.0}
\providecommand{\RbestBaseAcc}{0.0}
\providecommand{\RfidBetaTwo}{0}
\providecommand{\RfidBetaZero}{0}

\providecommand{\RoursAccLo}{0.0}
\providecommand{\RoursAccHi}{0.0}
\providecommand{\RpsnrTwo}{0.0}
\providecommand{\RpsnrZero}{0.0}

\providecommand{\RheroPSNR}{0.0}
\providecommand{\RftGapFrozen}{0}
\providecommand{\RftGapTuned}{0}
\providecommand{\RftOursTuned}{0}
\providecommand{\RftBaseTuned}{0}
\providecommand{\RcoupleScramble}{0.00}
\providecommand{\RdecoupleBeta}{0.00}
\providecommand{\RaccScrambleHi}{0.0}
\providecommand{\RaccScrambleLo}{0.0}
\providecommand{\RspecRelErr}{0.000}
\IfFileExists{numbers.tex}{\input{numbers.tex}}{}
\newcommand{\incgfx}[2]{\IfFileExists{#2}{\includegraphics[width=#1]{#2}}%
  {\fbox{\parbox[c][3cm][c]{#1}{\centering\small (figure pending)}}}}

\title{\vspace{-1.5em}Naturalness Predicts but Does Not Cause Transferability\\
in Image Encodings of Real-World Streams}
\author{Faruk Alpay\thanks{Corresponding author: \texttt{alpay@lightcap.ai}} \quad Barış Başaran\\[3pt]
  \small Department of Computer Engineering, Bahçeşehir University, Istanbul, Turkey\\[-1pt]
  \small \texttt{\{faruk.alpay, baris.basaran\}@bahcesehir.edu.tr}}
\date{}

\begin{document}
\maketitle

\begin{abstract}
\noindent
A common practice converts a one-dimensional signal into an image so that a vision
backbone pretrained on natural photographs can be reused for recognition. Such
encodings are usually evaluated by downstream accuracy, and the image itself is
rarely examined. We study, on real data, how the visual ``naturalness'' of an
encoded image relates to its transfer accuracy under a frozen backbone. We build
WorldStream, a corpus of \Rnseries{} heterogeneous current-value series collected
from key-free public APIs (weather, air quality, earthquakes, gold and oil,
equities, cryptocurrencies, foreign exchange, community attention and space
weather), and define a nine-way source-recognition task over \Rnwin{} temporally
split windows. Across seven image encodings and six frozen backbones, the
Fr\'echet distance of an encoding to natural images (FID) predicts its accuracy
well: the Spearman correlation is $\rho=\Rspear$. Two controlled interventions show
that the correlation is not causal in the spectrum. We build an invertible encoder
whose only adjustable component is a spectral exponent $\beta$
(power $\propto|f|^{-\beta}$); varying $\beta$ moves the image toward or away from the
natural-image manifold while leaving the encoded signal fixed. FID is lowest near the
natural value $\beta\approx 2$, but frozen accuracy stays flat and far below the
structured baselines (\RoursAcc\% against \RbestBaseAcc\%), and over this sweep FID
and accuracy are only weakly related (Pearson $\RdecoupleBeta$). A second intervention, phase
scrambling, holds the power spectrum exactly fixed while removing local structure;
now FID and accuracy fall together (Pearson $\RcoupleScramble$). The cross-encoding
correlation is therefore mediated by local structure, not by spectral naturalness:
FID predicts accuracy because the Inception network behind it reads the same local
structure the backbones read. Full fine-tuning does not close the gap
(\RftOursTuned\% against \RftBaseTuned\%), confirming that the deficit is structural.
The encoder is exactly invertible: the signal is recovered from the 8-bit image at
\RpsnrTwo~dB, so the same picture serves as a lossless record of the data, which we
use to render the world state of June~2026 as a shaded-relief landform that inverts
back to the numbers it depicts.
\end{abstract}

\section{Introduction}
A recurring step in applied pattern recognition is to turn a signal into a picture
and hand it to a convolutional or transformer backbone that was pretrained on
natural photographs. The Gramian Angular Field, the Markov Transition Field and the
recurrence plot~\cite{wang2015imaging,hatami2018recurrence} are standard encodings.
The same idea drives tabular-to-image methods such as DeepInsight, REFINED and
IGTD~\cite{sharma2019deepinsight,bazgir2020refined,zhu2021igtd}, and a growing body
of work that feeds encoded series to large vision and vision--language
models~\cite{survey2025vision}. These methods produce images that are visually
unlike natural scenes, yet they rely on backbones tuned to the statistics of
natural scenes.

This raises a concrete question. For a frozen backbone, does the accuracy of an
encoding depend on how close its images sit to the distribution of natural
photographs? Natural-image statistics give a precise notion of closeness. The
radially averaged power spectrum of photographs falls as $|f|^{-2}$, a
scale-invariant $1/f$ law that is stable across scenes and is also observed in
visual art~\cite{vanderschaaf1996,ruderman1994,torralba2003,koch2010}. If a frozen
backbone is tuned to that law, an encoding that respects it might transfer better.
We evaluate this hypothesis on a broad collection of real, current-value streams and
report two findings.

\begin{itemize}\setlength\itemsep{0.2em}
\item Across $7\times6$ encoding--backbone pairs, frozen accuracy is strongly
correlated with FID to natural images (Spearman $\rho=\Rspear$; up to $\RspearBest$
within a backbone). Naturalness is a good predictor of transfer.
\item The spectrum is not the cause. An invertible encoder whose spectral slope
$\beta$ is a single knob lets us change naturalness at fixed content. FID is lowest
near $\beta\approx 2$, but accuracy stays flat and far below the baselines, and FID
and accuracy are only weakly related over the sweep (Pearson $\RdecoupleBeta$).
\item Structure is the cause. Phase scrambling holds the power spectrum exactly fixed
and removes local structure; accuracy and FID then fall together (Pearson
$\RcoupleScramble$). The cross-encoding correlation is mediated by local structure,
which the FID network and the backbones both read.
\item Fine-tuning does not rescue it (\RftOursTuned\% against \RftBaseTuned\%), so the
deficit is structural rather than specific to frozen transfer.
\end{itemize}

\begin{figure}[H]
\centering
\incgfx{\linewidth}{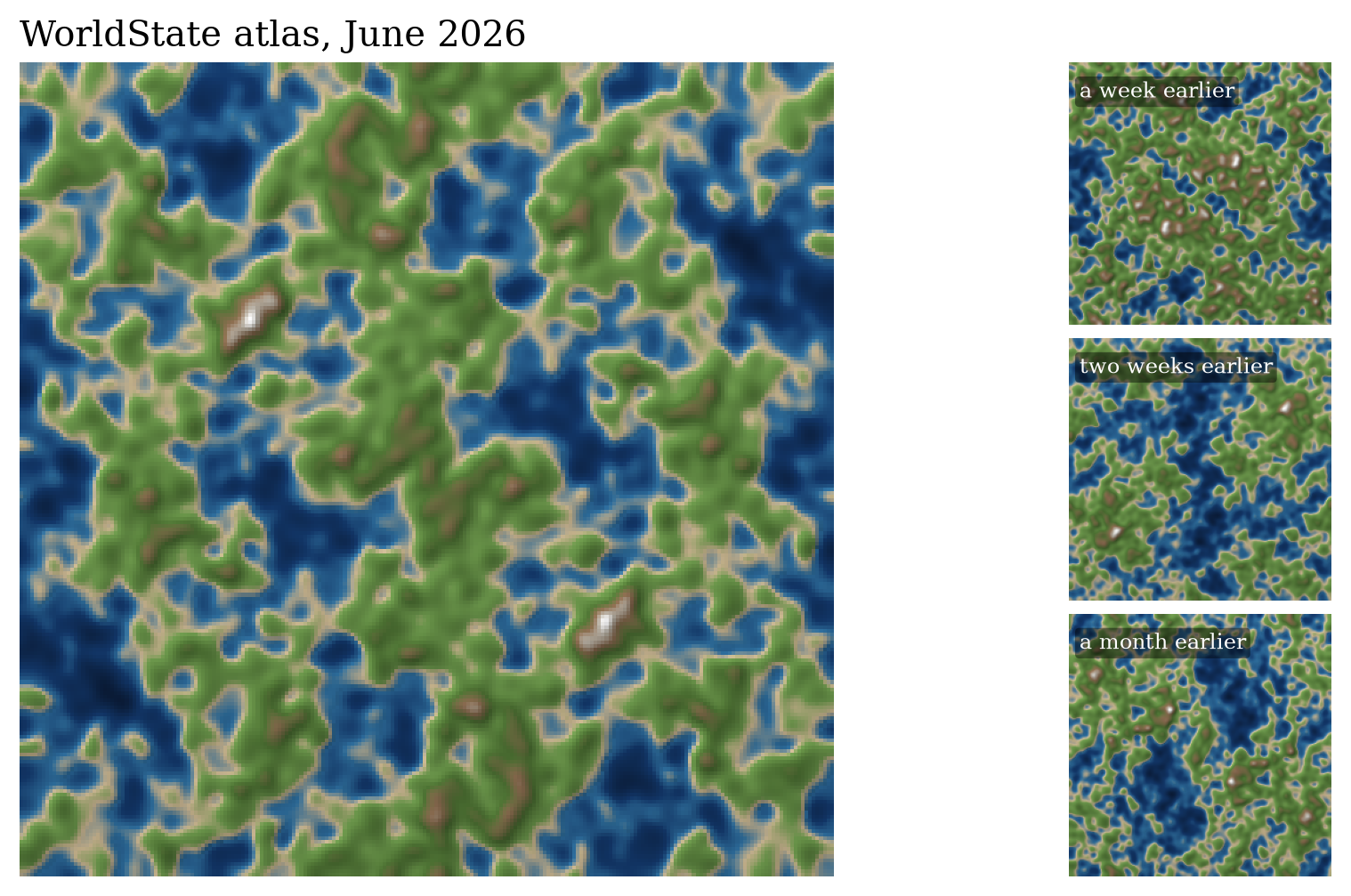}
\caption{A length-$L$ vector built from the most recent observations of twelve
streams (gold, oil, the S\&P~500, the VIX, Bitcoin, EUR/USD, Tokyo temperature,
Californian seismicity, Hacker-News volume, Wikipedia attention, solar-wind speed
and London air quality), rendered by the $\beta{=}2$ encoder as a shaded-relief
landform. The three smaller panels repeat the construction for earlier world
states. The image is not an illustration of the data; it is the data. The carrier
field inverts to the source vector at \RheroPSNR~dB (Proposition~\ref{prop:inv}), so
elevation, coastline and ridges are deterministic functions of the numbers.}
\label{fig:hero}
\end{figure}

The encoder used for the controlled comparison is invertible, so it also serves as a
lossless visualisation: a single image is at once a recognition input and an exact
record of the underlying signal, as illustrated in Figure~\ref{fig:hero}.

\section{Related work}
\textbf{Imaging a signal.} Encoding time series as images to use convolutional
networks goes back to the Gramian Angular and Markov Transition
Fields~\cite{wang2015imaging} and recurrence
plots~\cite{eckmann1987,hatami2018recurrence}. Recent surveys add line plots,
heatmaps and spectrograms~\cite{survey2025vision}. Tabular-to-image methods place
features on a grid to preserve neighbourhood structure
(DeepInsight~\cite{sharma2019deepinsight}, REFINED~\cite{bazgir2020refined},
IGTD~\cite{zhu2021igtd}). All of these optimise a downstream network. None measures
or targets proximity to the natural-image distribution, and invertibility is left
as a property that some encodings happen to have.

\textbf{Transfer and the domain gap.} Whether ImageNet accuracy predicts downstream
accuracy is a long-running question~\cite{kornblith2019,battle2023}. The gap between
a pretraining distribution and a target distribution is often estimated in a
pretrained feature space, for example with Fr\'echet distances~\cite{heusel2017fid}.
We apply this view to signal imaging: the target distribution is the set of encoded
pictures, and we relate its distance from natural images to transfer accuracy.

\textbf{Natural-image statistics.} The $1/f$ amplitude (that is, $1/f^2$ power)
spectrum of natural scenes is among the most robust regularities in vision
science~\cite{vanderschaaf1996,ruderman1994,torralba2003} and persists in human-made
imagery~\cite{koch2010}. We use it as the prior that defines the encoder's
naturalness knob.

\section{The WorldStream corpus}
\label{sec:data}
We collect heterogeneous current-value series from public, key-free endpoints and
group them into nine source families (Table~\ref{tab:corpus}): hourly weather and
air quality for sixteen cities (Open-Meteo archive); regional and global earthquake
activity (USGS); commodities including gold and crude oil, equity indices and single
names, major cryptocurrencies and foreign-exchange rates (Yahoo Finance, CoinGecko,
Frankfurter); community attention from Hacker News and Wikipedia pageviews; and
solar-wind telemetry (NOAA SWPC). Each series stores its source URL and fetch time,
and the manifest is shipped with the code.

From every series we take length-$L{=}192$ windows at stride $64$, discard
near-constant windows, and z-score each window so that absolute scale is removed and
the task depends on dynamics rather than magnitude. Windows are balanced across the
nine families and split in time: within each series the earliest $70\%$ of windows
form the training set and the latest $30\%$ the test set, with a one-window gap to
avoid overlap. This gives \Rntrain{} training and \Rntest{} test windows and exposes
the recogniser to genuine drift. The task is nine-way source-family recognition:
given a window's image, identify the family that produced it. Because windows are
scale-normalised, the usable cue is shape and temporal structure.

\input{corpus_table}

\section{An invertible spectral encoder}
\label{sec:encoder}
The controlled experiments of Section~\ref{sec:results} require an encoder whose power
spectrum can be varied continuously while the encoded signal is held fixed. The
following construction meets this requirement and is invertible, so that the same
image is also a lossless record of the data.

Let $x\in\mathbb{R}^{L}$ be a z-scored window and fix an $N\times N$ grid ($N{=}224$).
Let $\{(i_m,j_m)\}_{m=1}^{P}$ denote the $P$ lowest radial-frequency interior modes of
the real two-dimensional DFT (columns $1\le j\le N/2-1$, for which
$\mathrm{irfft2}\circ\mathrm{rfft2}$ is exact and the real and imaginary parts are
unconstrained), with radial frequencies $|f_m|$. Fix an orthonormal frame
$Q\in\mathbb{R}^{P\times L}$ with $Q^{\top}Q=I_L$ ($P{=}2048$) and an amplitude
envelope $a_m=|f_m|^{-\beta/2}$ normalised to unit norm. With $c=Qx$, the real field
\begin{equation}
H = \mathrm{irfft2}\!\Big(\textstyle\sum_m a_m c_m\,\delta_{(i_m,j_m)}\Big)
\end{equation}
is mapped to an image $I=C(g)$ through a fixed monotone tone map
$g=\tfrac12(\tanh(H/s)+1)\in(0,1)$ and a fixed injective colormap $C$.

Because $c=Qx$ distributes the signal across the $P$ modes and $a_m$ scales mode $m$,
the expected power of $H$ at frequency $|f_m|$ is $\propto a_m^2=|f_m|^{-\beta}$, so
the radially averaged spectrum of $H$ is $\propto|f|^{-\beta}$ by construction and
$\beta{=}2$ reproduces the $1/f^2$ law of natural images~\cite{vanderschaaf1996}. The
exponent $\beta$ is therefore a single parameter that moves the encoded image toward
or away from the natural-image manifold without altering the $L$ numbers it encodes.

\begin{proposition}[Exact inversion]\label{prop:inv}
For an unquantised carrier $g$, the maps $H=s\,\mathrm{arctanh}(2g-1)$ and
$x=Q^{\top}\big(\mathrm{Re}\,\mathrm{rfft2}(H)[i_m,j_m]/a_m\big)_m$ recover $x$ exactly,
since $Q^{\top}Q=I_L$ and the selected modes round-trip without a Hermitian
constraint. Under 8-bit quantisation the error is bounded: the reconstruction PSNR is
\RpsnrTwo~dB at $\beta{=}2$ and \RpsnrZero~dB at $\beta{=}0$
(Section~\ref{sec:results}).
\end{proposition}

\section{Experimental setup}
\label{sec:setup}
The baseline encodings are the line plot, the spectrogram, the Gramian Angular
Summation and Difference Fields, the Markov Transition Field and the recurrence plot;
the proposed encoder is evaluated over $\beta\in\{0,0.5,1,1.5,2,2.5,3,4\}$. A single
colormap is shared by every encoding, so that spatial structure is the only quantity
that differs across methods and the spectral effect is isolated from colour;
Figure~\ref{fig:gallery} shows one window under all eight encodings. We use six frozen
pretrained backbones as fixed feature extractors (ResNet-50, ViT-B/16, CLIP ViT-B/16,
DINOv2 ViT-B/14, DINOv2 ViT-L/14 and ConvNeXt-L) and fit a logistic-regression linear
probe on standardised features, reporting top-1 accuracy and macro-F1 on the temporal
test split with bootstrap $95\%$ confidence intervals. The Fr\'echet Inception
Distance (FID) is computed between $2000$ encoded training images and $2000$ natural
images drawn from Imagenette, an ImageNet subset. The fine-tuning study instead trains
ResNet-50 and ViT-B/16 end to end on the encoded images. All computation runs on a
single NVIDIA RTX~5090 (32\,GB, Blackwell) with PyTorch and CUDA~12.8.

\begin{figure}[H]
\centering
\incgfx{\linewidth}{fig_gallery.png}
\caption{A single window under the eight encodings (left to right: line plot, GASF,
MTF, recurrence plot, spectrogram, and the proposed encoder at $\beta{=}0,2,4$). The
encodings share one colormap, so the panels differ only in spatial structure.}
\label{fig:gallery}
\end{figure}

\section{Results}
\label{sec:results}\label{sec:law}
\begin{figure}[H]
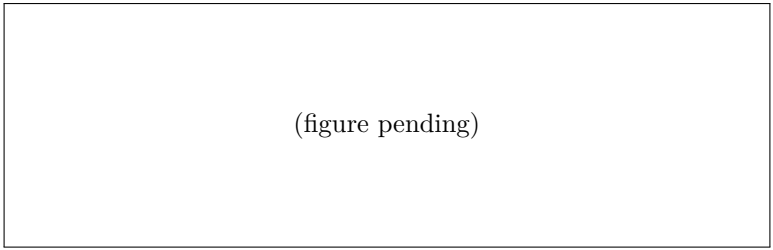

\centering
\incgfx{0.6\linewidth}{fig_scatter.pdf}
\caption{Frozen linear-probe accuracy against FID to natural images, one point per
encoding and backbone. Lower FID accompanies higher accuracy, with Spearman
$\rho=\Rspear$.}
\label{fig:scatter}
\end{figure}

Frozen transfer accuracy is strongly and monotonically correlated with FID to natural
images across the $7\times6$ encoding--backbone pairs, with Spearman $\rho=\Rspear$
overall and as strong as $\RspearBest$ within a single backbone
(Figure~\ref{fig:scatter}); Table~\ref{tab:main} reports the full grid. The
natural-looking baselines, namely the Gramian and Markov fields and the recurrence
plot, reach \RbestBaseAcc\% accuracy, whereas the proposed spectral encoder, which
distributes each sample across global Fourier modes, remains near the lower end of
both axes. On its own this correlation invites the conclusion that a more natural
encoding transfers better. The remainder of this section shows, through two controlled
interventions, that the correlation is mediated by local image structure rather than
by spectral naturalness.

\input{main_table}

The first intervention varies the spectral exponent $\beta$, which changes the
radially averaged power spectrum of the proposed encoder while leaving the $L$ encoded
numbers unchanged. As $\beta$ increases from $0$, the FID of the encoder falls from
\RfidBetaZero{} for a white, high-frequency field to a minimum of \RfidBetaTwo{} near
the natural value $\beta\approx 2$, and rises again as the field over-smooths; the
power spectrum at $\beta{=}2$ follows the $1/f^2$ law of natural photographs over more
than a decade of spatial frequency. Accuracy, however, does not track FID over this
range. It stays within \RoursAccLo--\RoursAccHi\%, far below the \RbestBaseAcc\% of the
structured baselines, with no maximum at $\beta\approx 2$, and FID and accuracy are
only weakly correlated across the sweep (Pearson $\RdecoupleBeta$). Increasing the
naturalness of the image at fixed content therefore produces no measurable gain in
transfer.

\begin{figure}[H]
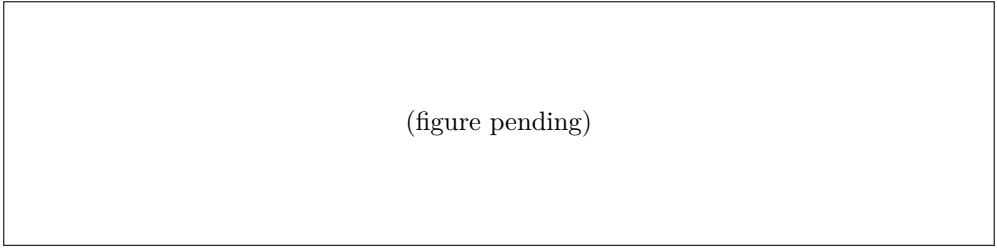

\centering
\incgfx{0.78\linewidth}{fig_structure.pdf}
\caption{Phase scrambling of a recurrence-plot encoding at three levels $f$
(illustrative example). The three panels share an identical power spectrum; as $f$
grows the local structure is destroyed. On the corpus this operation couples FID and
frozen accuracy at Pearson $\RcoupleScramble$ over the sweep.}
\label{fig:structure}
\end{figure}

The second intervention holds the spectrum fixed and instead removes local structure.
Randomising a fraction $f$ of the Fourier phases of an image leaves the magnitude of
every frequency, and hence the entire power spectrum, unchanged, while progressively
destroying spatial structure (Figure~\ref{fig:structure}). Applied to the two most
readable baselines, recurrence and Gramian, the procedure preserves the power spectrum
throughout (mean relative change \RspecRelErr) yet degrades FID and accuracy together:
recurrence accuracy falls from \RaccScrambleHi\% at $f{=}0$ to \RaccScrambleLo\% at
$f{=}1$, and FID and accuracy are strongly correlated over the sweep (Pearson
$\RcoupleScramble$). With the spectrum held constant, removing structure moves both
the predictor and the target.

Together the interventions identify the property that the correlation reflects.
Varying the spectrum at fixed structure leaves FID and accuracy only weakly related
(Pearson $\RdecoupleBeta$), whereas varying the structure at fixed spectrum couples
them strongly (Pearson $\RcoupleScramble$). The cross-encoding correlation of
Figure~\ref{fig:scatter} is thus mediated by local structure rather than by the
spectrum: FID predicts accuracy because the Inception network that defines it responds
to the same local structure that the backbones exploit. A Gramian or recurrence image
arranges pairwise relations between samples in the image plane, where a backbone's
local-to-intermediate receptive fields can read them, whereas the global Fourier
synthesis of the proposed encoder spreads each sample across the whole image and
leaves no local arrangement to read. The encodings that look natural are also the ones
that are locally structured, and it is the structure, not the spectrum, that the
correlation tracks.

End-to-end fine-tuning does not remove the gap (Table~\ref{tab:ft}). It raises the
structured baselines but barely changes the spectral encoder, which remains at
\RftOursTuned\% against \RftBaseTuned\% and above for the baselines; the spread between
the most and least natural encoding is \RftGapFrozen{} accuracy points under the frozen
probe and \RftGapTuned{} points after tuning. A backbone with full gradient access
still cannot read a globally mixed encoding, so the deficit is structural rather than
specific to frozen transfer, and the convolutional ResNet recovers less than the
attention-based ViT, consistent with the missing property being local arrangement.

\IfFileExists{ft_table.tex}{\input{ft_table}}{}

The encoder is finally lossless to within quantisation: the 8-bit reconstruction PSNR
ranges from \RpsnrZero~dB at $\beta{=}0$ to \RpsnrTwo~dB at $\beta{=}2$, so the image
that is fed to a backbone also recovers the source signal. This makes the encoder
useful as a faithful visualisation rather than as a recogniser. Figure~\ref{fig:hero}
renders a real world-state vector at $\beta{=}2$ as a shaded-relief landform that
inverts to the source numbers at \RheroPSNR~dB.

\section{Discussion}
FID to natural images is a strong and inexpensive predictor of frozen transfer
accuracy across encodings, but it is not a control variable: increasing the
naturalness of an image at fixed content does not improve transfer, and fine-tuning
does not close the gap, which locates the spectral encoder's weakness in its absence
of local structure. Several caveats bound the scope of this conclusion. The proposed
encoder is content-agnostic by design, since it distributes the signal globally in
order to isolate the spectrum, and is consequently a weak recogniser; we draw no
conclusion about encoders that combine a spectral prior with an explicit local layout,
which we expect to behave differently. FID is one proxy for naturalness, computed in
Inception feature space against a single reference set, and other manifold distances
may shift the constants, though we do not expect them to alter the separation between
the two interventions. We also attempted to learn the amplitude envelope by gradient
descent so as to match the Inception statistics of natural images; on this
band-limited encoder the optimisation moved energy toward the highest available
frequencies rather than reproducing the $1/f^2$ roll-off, which again indicates that
the property relevant to transfer is local edge content rather than the global
spectral slope.

\section{Conclusion}
A natural-looking signal-to-image encoding tends to transfer better under a frozen
backbone, but the relationship is predictive rather than causal. Two controlled
interventions separate the candidate causes: varying the spectrum at fixed structure
leaves accuracy unchanged, whereas varying the structure at fixed spectrum moves
accuracy and FID together. The cross-encoding correlation is therefore mediated by
local image structure, which both the Inception network behind FID and the vision
backbones read; spectral naturalness alone does not determine transfer, and
fine-tuning does not compensate for an encoding that lacks local structure. The encoder
used for these experiments is invertible, so the same image that is fed to a backbone
also recovers the underlying signal. We release the WorldStream data, the code and the
data manifest to support replication.

{\small
\bibliographystyle{plain}
\bibliography{refs}}

\end{document}

%% file: numbers.tex
\renewcommand{\Rspear}{-0.72}
\renewcommand{\RspearBest}{-0.79}
\renewcommand{\Rntrain}{2322}
\renewcommand{\Rntest}{821}

\renewcommand{\Rnseries}{299}
\renewcommand{\Rnwin}{3143}

\renewcommand{\RoursAcc}{19.2}
\renewcommand{\RoursAccLo}{13.9}
\renewcommand{\RoursAccHi}{23.6}
\renewcommand{\RbestBaseAcc}{73.0}
\renewcommand{\RfidBetaTwo}{348}
\renewcommand{\RfidBetaZero}{387}

\renewcommand{\RpsnrTwo}{72.9}
\renewcommand{\RpsnrZero}{78.3}

\renewcommand{\RheroPSNR}{74.5}
\renewcommand{\RftGapFrozen}{53}
\renewcommand{\RftGapTuned}{55}
\renewcommand{\RftOursTuned}{27}
\renewcommand{\RftBaseTuned}{67}
\renewcommand{\RcoupleScramble}{-0.89}
\renewcommand{\RdecoupleBeta}{-0.32}
\renewcommand{\RaccScrambleHi}{67.1}
\renewcommand{\RaccScrambleLo}{56.5}
\renewcommand{\RspecRelErr}{0.019}

%% file: corpus_table.tex
\begin{table}[H]\centering\small
\caption{The WorldStream corpus: nine source families, 299 real series, 3143 balanced windows.}
\label{tab:corpus}
\begin{tabular}{llcc}
\toprule
Family & Sources & Series & Win.\,(tr/te) \\
\midrule
airquality & Open-Meteo AQ & 60 & 259/111 \\
commodity & Yahoo (gold, oil) & 10 & 250/60 \\
crypto & Yahoo, CoinGecko & 34 & 259/86 \\
fx & Yahoo, Frankfurter & 26 & 259/72 \\
market & Yahoo (indices, equities) & 30 & 259/111 \\
seismic & USGS & 19 & 259/111 \\
spaceweather & NOAA SWPC & 9 & 259/111 \\
weather & Open-Meteo & 80 & 259/111 \\
webactivity & Hacker News, Wikipedia & 31 & 259/48 \\
\midrule
\textbf{total} & & 299 & 2322/821 \\
\bottomrule
\end{tabular}
\end{table}

%% file: main_table.tex
\begin{table}[H]\centering\footnotesize\setlength{\tabcolsep}{4.5pt}
\caption{Nine-way source-family recognition: frozen linear-probe accuracy (\%) and FID-to-natural for every encoding and backbone. Best per column in bold; the $\beta{=}2$ row is the natural-image prior. Lower FID is more natural.}
\label{tab:main}
\begin{tabular}{lccccccc}
\toprule
Encoding & FID$\downarrow$ & CLIP-B & ConvNeXt-L & DINOv2-B & DINOv2-L & ResNet-50 & ViT-B/16 \\
\midrule
line plot & 269 & \textbf{73.0} & \textbf{72.0} & \textbf{70.9} & \textbf{72.4} & 64.3 & \textbf{72.5} \\
spectrogram & 294 & 54.2 & 58.6 & 56.3 & 57.6 & 50.3 & 54.8 \\
GASF & 268 & 63.5 & 68.1 & 66.1 & 66.0 & 62.7 & 65.2 \\
GADF & 268 & 63.3 & 69.5 & 67.1 & 67.6 & 59.0 & 65.5 \\
MTF & 257 & 57.5 & 61.3 & 56.2 & 57.2 & 56.4 & 57.4 \\
recurrence & 289 & 63.9 & 69.3 & 67.6 & 66.4 & \textbf{66.7} & 65.8 \\
\midrule
ours $\beta{=}0$ & 387 & 19.4 & 21.0 & 21.0 & 20.2 & 17.9 & 21.2 \\
ours $\beta{=}0.5$ & 382 & 21.2 & 20.8 & 23.0 & 20.1 & 16.6 & 21.7 \\
ours $\beta{=}1$ & 372 & 20.5 & 20.5 & 20.7 & 17.7 & 15.0 & 19.9 \\
ours $\beta{=}1.5$ & 360 & 17.3 & 20.1 & 19.2 & 16.6 & 16.2 & 18.5 \\
ours $\beta{=}2$ & 348 & 18.3 & 20.8 & 19.7 & 16.6 & 13.9 & 20.3 \\
ours $\beta{=}2.5$ & 344 & 22.4 & 20.7 & 19.6 & 17.4 & 16.4 & 19.9 \\
ours $\beta{=}3$ & 366 & 23.6 & 22.3 & 19.6 & 17.1 & 17.5 & 19.6 \\
ours $\beta{=}4$ & 444 & 17.3 & 18.1 & 17.4 & 19.1 & 16.2 & 19.4 \\
\bottomrule
\end{tabular}
\end{table}

%% file: ft_table.tex
\begin{table}[H]\centering\small
\caption{Frozen linear probe vs.\ full fine-tuning (test accuracy, \%). Fine-tuning lifts every encoding but does not close the gap: the spectral encoder stays far below the structured baselines even end-to-end, so the deficit is structural, not a frozen-transfer artefact.}
\label{tab:ft}
\begin{tabular}{lcccc}
\toprule
 & \multicolumn{2}{c}{ResNet-50} & \multicolumn{2}{c}{ViT-B/16} \\
\cmidrule(lr){2-3}\cmidrule(lr){4-5}
Encoding & frozen & tuned & frozen & tuned \\
\midrule
line plot & 64.3 & 67.5 & 72.5 & 75.6 \\
GASF & 62.7 & 70.4 & 65.2 & 73.7 \\
recurrence & 66.7 & 71.0 & 65.8 & 75.2 \\
ours $\beta{=}0$ & 17.9 & 15.6 & 21.2 & 26.1 \\
ours $\beta{=}2$ & 13.9 & 15.8 & 20.3 & 25.7 \\
ours $\beta{=}4$ & 16.2 & 16.3 & 19.4 & 26.6 \\
\bottomrule
\end{tabular}
\end{table}

%% file: main.bbl
\begin{thebibliography}{10}

\bibitem{bazgir2020refined}
Omid Bazgir, Ruibo Zhang, Saugato~Rahman Dhruba, Raziur Rahman, Souparno Ghosh,
  and Ranadip Pal.
\newblock Representation of features as images with neighborhood dependencies
  for compatibility with convolutional neural networks.
\newblock {\em Nature Communications}, 11(1):4391, 2020.

\bibitem{eckmann1987}
Jean-Pierre Eckmann, S~Oliffson Kamphorst, and David Ruelle.
\newblock Recurrence plots of dynamical systems.
\newblock {\em Europhysics Letters}, 4(9):973--977, 1987.

\bibitem{battle2023}
Micah Goldblum, Hossein Souri, Renkun Ni, Manli Shu, Viraj Prabhu, Gowthami
  Somepalli, Prithvijit Chattopadhyay, Mark Ibrahim, Adrien Bardes, Judy
  Hoffman, et~al.
\newblock Battle of the backbones: A large-scale comparison of pretrained
  models across computer vision tasks.
\newblock In {\em Advances in Neural Information Processing Systems (NeurIPS)},
  2023.

\bibitem{hatami2018recurrence}
Nima Hatami, Yann Gavet, and Johan Debayle.
\newblock Classification of time-series images using deep convolutional neural
  networks.
\newblock In {\em Tenth International Conference on Machine Vision (ICMV
  2017)}, volume 10696, pages 242--249. SPIE, 2018.

\bibitem{heusel2017fid}
Martin Heusel, Hubert Ramsauer, Thomas Unterthiner, Bernhard Nessler, and Sepp
  Hochreiter.
\newblock Gans trained by a two time-scale update rule converge to a local nash
  equilibrium.
\newblock In {\em Advances in Neural Information Processing Systems (NeurIPS)},
  2017.

\bibitem{koch2010}
Marcel Koch, Joachim Denzler, and Christoph Redies.
\newblock A {1/f$^2$} characteristic and isotropy in the {Fourier} power
  spectra of visual art, cartoons, comics, mangas, and different categories of
  photographs.
\newblock {\em PLoS ONE}, 5(8):e12268, 2010.

\bibitem{kornblith2019}
Simon Kornblith, Jonathon Shlens, and Quoc~V Le.
\newblock Do better imagenet models transfer better?
\newblock In {\em Proceedings of the IEEE/CVF Conference on Computer Vision and
  Pattern Recognition (CVPR)}, pages 2661--2671, 2019.

\bibitem{survey2025vision}
Jingchao Ni, Ziming Zhao, ChengAo Shen, Hanghang Tong, Dongjin Song, Wei Cheng,
  Dongsheng Luo, and Haifeng Chen.
\newblock Harnessing vision models for time series analysis: A survey.
\newblock {\em arXiv preprint arXiv:2502.08869}, 2025.

\bibitem{ruderman1994}
Daniel~L Ruderman and William Bialek.
\newblock Statistics of natural images: Scaling in the woods.
\newblock {\em Physical Review Letters}, 73(6):814--817, 1994.

\bibitem{sharma2019deepinsight}
Alok Sharma, Edwin Vans, Daichi Shigemizu, Keith~A Boroevich, and Tatsuhiko
  Tsunoda.
\newblock Deepinsight: A methodology to transform a non-image data to an image
  for convolution neural network architecture.
\newblock {\em Scientific Reports}, 9(1):11399, 2019.

\bibitem{torralba2003}
Antonio Torralba and Aude Oliva.
\newblock Statistics of natural image categories.
\newblock {\em Network: Computation in Neural Systems}, 14(3):391--412, 2003.

\bibitem{vanderschaaf1996}
A~van~der Schaaf and J~H van Hateren.
\newblock Modelling the power spectra of natural images: statistics and
  information.
\newblock {\em Vision Research}, 36(17):2759--2770, 1996.

\bibitem{wang2015imaging}
Zhiguang Wang and Tim Oates.
\newblock Imaging time-series to improve classification and imputation.
\newblock In {\em Proceedings of the 24th International Joint Conference on
  Artificial Intelligence (IJCAI)}, pages 3939--3945, 2015.

\bibitem{zhu2021igtd}
Yitan Zhu, Thomas Brettin, Fangfang Xia, Alexander Partin, Maulik Shukla,
  Hyunseung Yoo, Yvonne~A Evrard, James~H Doroshow, and Rick~L Stevens.
\newblock Converting tabular data into images for deep learning with
  convolutional neural networks.
\newblock {\em Scientific Reports}, 11(1):11325, 2021.

\end{thebibliography}
